%% file: custom.tex
\pdfoutput=1

\documentclass[11pt]{article}

\usepackage[final]{acl}

\usepackage{times}
\usepackage{latexsym}
\usepackage{amssymb}
\usepackage[T1]{fontenc}

\usepackage[utf8]{inputenc}
\usepackage{amsmath}
\usepackage{microtype}

\usepackage{inconsolata}
\usepackage{algorithm} 
\usepackage{algpseudocode} 
\usepackage{graphicx}
\usepackage{multirow}
\usepackage{booktabs}
\usepackage{subcaption}
\usepackage[normalem]{ulem}
\useunder{\uline}{\ul}{}
\usepackage{float}
\title{Unshackling Context Length: An Efficient Selective Attention Approach through Query-Key Compression}
\author{
	Haoyu Wang\textsuperscript{\rm 1}$^*$, Tong Teng \textsuperscript{\rm 1}\thanks{Equal Contribution.}, Tianyu Guo\textsuperscript{\rm 1}, An Xiao\textsuperscript{\rm 1}, Duyu Tang\textsuperscript{\rm 2},   \\
	\textbf{Hanting Chen\textsuperscript{\rm 1}, Yunhe Wang\textsuperscript{\rm 1}\thanks{Corresponding Author.}} \\
	\textsuperscript{\rm 1}Huawei Noah’s Ark Lab ~~~\textsuperscript{\rm 2}Huawei CBG  \\
	\texttt{\small \{wanghaoyu50, tengtong1, tianyu.guo, an.xiao, tangduyu, chenhanting, yunhe.wang\}@huawei.com}
}

\begin{document}
\maketitle
\begin{abstract}

Handling long-context sequences efficiently remains a significant challenge in large language models (LLMs). Existing methods for token selection in sequence extrapolation either employ a permanent eviction strategy or select tokens by chunk, which may lead to the loss of critical information. We propose Efficient Selective Attention (ESA), a novel approach that extends context length by efficiently selecting the most critical tokens at the token level to compute attention. ESA reduces the computational complexity of token selection by compressing query and key vectors into lower-dimensional representations. We evaluate ESA on long sequence benchmarks with maximum lengths up to 256k using open-source LLMs with context lengths of 8k and 32k. ESA outperforms other selective attention methods, especially in tasks requiring the retrieval of multiple pieces of information, achieving comparable performance to full-attention extrapolation methods across various tasks, with superior results in certain tasks. 

\end{abstract}

\section{Introduction}
\input{intro_v2}

\section{Related Work}
\paragraph{Position Extrapolation.}
Following the introduction of RoPE \citep{su2024roformer}, great efforts have been imposed to extend the context length by modifying the position embeddings (PE).
Position interpolation  \cite{chen2023extending, kaiokendev} extends the context length by interpolating positional indices within the constraints of pre-training. 
The NTK-aware method \citep{bloc97ntk,Rozire2023CodeLO,Liu2023ScalingLO} introduces a nonlinear interpolation strategy by increasing the base parameter of RoPE. YaRN \citep{peng2023yarn} proposes a method for frequency interpolation of RoPE dimensions, where higher frequency dimensions are extrapolated, and lower frequency dimensions are interpolated. Further improvements \citep{chen2023clex,Ding2024LongRoPEEL} exploit the dynamics in position extrapolation.
Another group of work redesigns the relative position matrix to overcome the OOD issue \citep{JianlinSu,Jin2024LLMML,An2024TrainingFreeLS}. 
These methods extend the context length but still compute the full attention matrix for inference thus fail to reduce the computational cost. To achieve better performance, some require fine-tuning with a certain amount of long-context data.

\paragraph{Selective Attention.}
Selective attention mechanisms aim to mitigate the computational cost of processing long sequences by selecting only the most relevant tokens for attention computation. Approaches like Longformer \citep{Beltagy2020LongformerTL} and BigBird \citep{Zaheer2020BigBT} use fixed or adaptive sparse patterns, while \citealp{han2023lm, xiao2023efficient} introduce $\Lambda$-shaped windows that evict middle tokens. Although these methods lower costs, they often compromise global context understanding due to restricted attention over all tokens.
Some methods aim at compressing the KV cache, usually perform token selection only in the decoding stage \citep{zhang2024h2o,liu2024scissorhands,ge2023model} or permanently evicting certain tokens \citep{xiao2023efficient, han2023lm, li2024snapkv}. While effective, they may lose critical contextual information.
As for chunk-based methods, \citealp{xiao2024infllm} uses an efficient contextual memory mechanism to select the most relevant chunks for computing attention, and \citealp{lu2024longheads} selects the most relevant chunks for each head separately considering the variance among different heads.
Unlimiformer and its adaptation \citep{bertsch2024unlimiformer,ahrabian2024adaptation} segment input during the pre-filling stage, using external memory blocks, but remain computationally expensive and require additional training or architecture modifications.
In contrast, our method performs efficient token-level selection in both prefilling and decoding stages, without discarding tokens permanently.

\section{Method}
\begin{figure*}
    \centering
    \begin{subfigure}[b]{\textwidth}
    \centering
    \includegraphics[trim=0cm 14.5cm 17.1cm 0cm, clip, width=\textwidth]{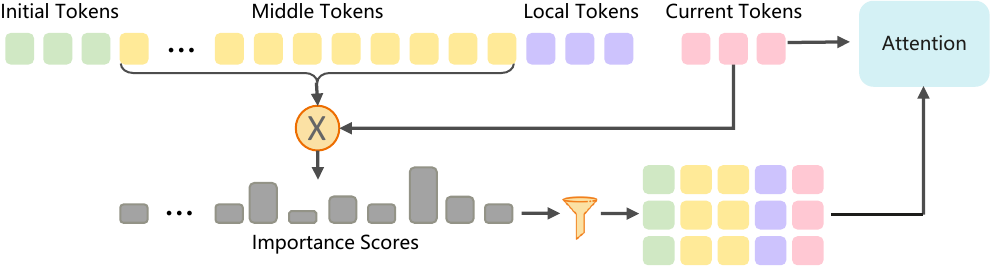}
    \vspace{-20pt}
    \caption{}
    \end{subfigure}

    \begin{subfigure}[b]{0.45\textwidth}
    \includegraphics[trim=0cm 14.6cm 24.7cm 0cm, clip, width=\textwidth]{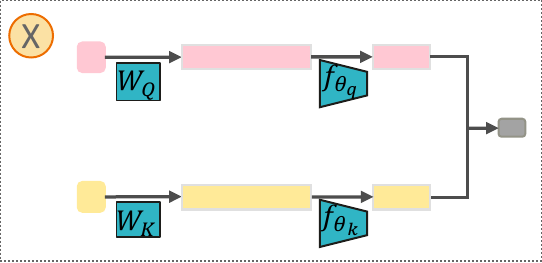}
    \caption{}
    \end{subfigure}
    \hfill
    \begin{subfigure}[b]{0.45\textwidth}
    \includegraphics[trim=0cm 14.6cm 24.7cm 0cm, clip, width=\textwidth]{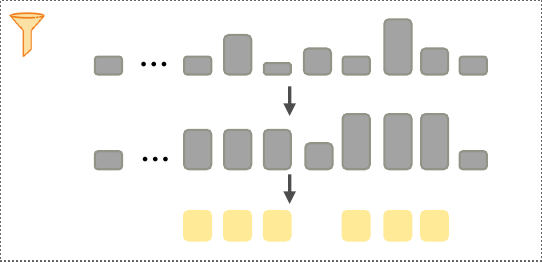}
    \caption{}
    \end{subfigure}
    \caption {(a) In long-context scenarios, the number of middle tokens occupies the majority, while the lengths of the other three parts of tokens are fixed. The importance scores between current tokens and middle tokens are utilized to select the top-k middle tokens. The selected tokens replace the middle tokens for computing attention. (b) The queries from current tokens and keys from middle tokens are compressed into smaller tensors through a linear layer respectively. The dot product of the compressed queries and keys serves as the importance scores. (c) The priority of a middle token being selected is determined by the maximum importance score among itself and several surrounding tokens.}
    \label{fig:illust}
\end{figure*}

Conventionally, each token attends to all preceding tokens to compute the attention output during LLM inference, leading to significant computational overhead when processing long-context sequences. 
We introduce ESA, which effectively identifies a subset of the most crucial tokens for attention computation, thereby improving the efficiency of the model.

\subsection{Token Selection}
Following previous work \citep{xiao2023efficient, han2023lm, xiao2024infllm} on extending the context window of LLMs, the preceding tokens ($P$) are segmented into three parts, as depicted in Figure \ref{fig:illust}: initial tokens ($I$), middle tokens ($M$), and local tokens ($L$) with respective lengths denoted as $l_I$, $l_M$, and $l_L$. With a massive number of preceding tokens, $l_M$ is considerably larger than $l_I$ and $l_L$.
The initial tokens and local tokens with fixed length will always be selected due to their significance, as highlighted by \citet{xiao2023efficient} and \citet{han2023lm}. We then apply ESA to select the top-$k$ critical tokens $M_k$ from the middle tokens $M$.
Specifically, LLM inference consists of two stages: prefilling and decoding. In the prefilling stage, all the input tokens are encoded, and the keys as well as values are computed and cached. We adopt the chunked-prefill method \citep{agrawal2023sarathi}.
Compared to the original prefilling method, chunked-prefill approach effectively reduces memory usage. In the decoding stage, the model generates tokens one by one autoregressively. 
Denote the \emph{current tokens} (i.e., the current input chunk being processed or the most recently decoded token) as $C$, the attention output computed in each step between the current tokens and the selected tokens is defined as:
\begin{equation} \label{attention_eq}
	\mathbf{O} = \text{Attn}(\mathbf{Q}_C,\mathbf{K}_{[I,M_k,L,C]}, \mathbf{V}_{[I,M_k,L,C]})
\end{equation}
where, $\mathbf{Q}_C$ represents the queries of the current tokens, $\mathbf{K}_{[I,M_k,L,C]}$ and $\mathbf{V}_{[I,M_k,L,C]}$ respectively denote the concatenated keys and values of the selected and current tokens.

%
\paragraph{Importance Score.} 
Successfully identifying the most crucial tokens requires a function to precisely measure the importance of each token in $M$ w.r.t $C$. Let $m \in M$ be a specific token, we denote the importance score of $m$ as $F_s(m; C)$.
Given a predefined number $k < =l_M$, $k$ tokens with the highest scores will be selected, as formalized bellow:
\begin{equation}
	M_k = \arg \text{topK}(M; F_s, C, k)
\end{equation}
For an individual token $c \in C$, the degree to which it attends to a preceding token $m$ can be determined as following:
\begin{equation}
	f_s(m; c) = \sum_{h=1}^{H}\mathbf{q}_{h,c}\mathbf{k}_{h,m}^{T}
	\label{dot-p}
\end{equation}
where, $\mathbf{q}_{h,c},\mathbf{k}_{h, m} \in \mathbb{R}^d$ denote the query of $c$ and the key of $m$, respectively, for the $h$-th head. All of the $H$ $d$-dimensional attention heads are incorporated, and the selected tokens are shared across all the heads in a certain layer. This score can be used directly in the decoding stage for token selection as $C$ consists of only one token (i.e., \( l_C = 1 \)). While at the prefilling stage, $l_C$ is the chunk size,
and every token in $C$ contributes to the score $F_s(m; C)$.
To derive a unified form of importance score, we first regularize the score w.r.t each $c$ and then take the maximum value across all tokens in $C$.
Eventually, the importance score is formulated as
\begin{equation}
	F_s(m; C) = \max_{c \in C} \left(f_s(m;c) - \max_{m^{\prime} \in M} f_s(m^{\prime};c)\right)
        \label{F_s_1}
\end{equation}
1. The expression \(f_s(m;c) - \max_{m^{\prime} \in M} f_s(m^{\prime};c)\) indicates that each token in \(C\) is constrained relative to the maximum score in \(M\), preventing any individual token in \(C\) from exerting a dominating influence on the selection process.
2. In the prefilling stage, our goal is to ensure that high-scoring tokens in \(M\) are not overlooked. Therefore, we select the highest score of each token in \(C\) to represent the score in \(M\) by applying \(\max_{c \in C}\).


\paragraph{Efficient Selection through Query-Key Compression.} 
The aforementioned scoring method of employing dot product across a considerable number tokens is computational expensive. To achieve efficient selection, we perform dimensionality reduction on keys and queries.
The right-hand side of Equation~\ref{dot-p} is equivalent to concatenating $H$ heads followed by performing dot product. That is, $ f_s(m;c)=\mathbf{q}_{c}\mathbf{k}_{m}^{T}$ where $\mathbf{q}_{c} = [\mathbf{q}_{1,c};\mathbf{q}_{2,c};... ;\mathbf{q}_{H,c}]$, $\mathbf{k}_{m} = [\mathbf{k}_{1,m};\mathbf{k}_{2,m};... ;\mathbf{k}_{H,m}]$, and
$\mathbf{q}_{c},\mathbf{k}_{ m} \in \mathbb{R}^{d_{H}}$, $d_{H}=H \times d$.
Denote the dimensionality reduction on queries and keys as follows:
\begin{equation} \label{reduce_dim_formula}
\begin{aligned}
\mathbf{q}_{c}^{\prime} &\triangleq f_{\theta_q}(\mathbf{q}_{c}),\\
\mathbf{k}_{m}^{\prime} &\triangleq f_{\theta_k}(\mathbf{k}_{m}), 
\end{aligned}
\end{equation}
where, $\mathbf{q}_{c}^{\prime}, \mathbf{k}_{m}^{\prime} \in \mathbb{R}^{d^{\prime}}$, $d^{\prime} < d_{H}$. 
The dimension-reduced representation $\mathbf{k}_{m}^{\prime}$ will be cached and reused during the subsequent computation steps.
With the lower-dimensional surrogates of queries and keys, the importance score is approximated with
\begin{equation} \label{reduce_dim_dot}
    f_s(m; c) \approx \mathbf{q}^\prime_{c}\mathbf{k}^{\prime T}_{m}
\end{equation}
The computational cost is therefore reduced compared with using the full-dimensional queries and keys. 
To maintain accuracy, the lower-dimensional representations should retrain the token order as faithfully as possible.
To this end, we perform a one-time offline procedure to learn $f_{\theta_q}$ and $f_{\theta_k}$ in each layer by minimizing the discrepancy between the importance scores before and after dimensionality reduction, formally: 
\begin{equation} \label{train_theta}
\begin{aligned}
\min_{\theta_q, \theta_k}  \sum_{c \in C, m \in M} \left\lVert \mathbf{q}_{c} \mathbf{k}_{m}^{T} - f_{\theta_q}(\mathbf{q}_{c}) f_{\theta_k}(\mathbf{k}_{m})^{T} \right\rVert_{2}^{2}
\end{aligned}
\end{equation}
We model $f_{\theta_q}$ and $f_{\theta_k}$ jointly, where each is a linear layer that projects a high-dimensional input to a low-dimensional output.
In preparation of the training data for the neural networks, we first perform token selection with full-dimensional queries and keys with a calibration dataset. All the queries and keys calculated during the process are saved.
Subsequently, we use the saved queries and keys to train \(f_{\theta_q}\) and \(f_{\theta_k}\) for each layer.
The learnt $f_{\theta_q}$ and $f_{\theta_k}$ will be utilized in our ESA to compress queries and keys with dimensionality reduction.
Since the low-dimensional keys are cached and reused, the additional computational load introduced by Equation~\ref{reduce_dim_formula} is marginal compared to the reduction achieved by Equation~\ref{reduce_dim_dot}. We conducted a quantitative analysis of efficiency in Section \ref{Complexity_Analysis}.
\paragraph{Proximity Influence.}
Considering proximity influence, if a token in $M$ has a high importance score, it affects the scores of the surrounding tokens. We propose adaptive selection which is achieved by updating the importance score of a token based on its neighbors. Specifically, the score of the $j$-th token, where $j\in [l_I,l_I+l_M-1]$, is determined as the maximum score within a window of size $ \epsilon $ surrounding it. The importance score of the $j$-th token, computed using the low-dimensional query and key, is denoted by $ s_j $. The updated score is given by 
\begin{equation} \label{proximity_influence_eq}
    s_j^\prime = \max_{w=\max(j-\epsilon, l_I)}^{\min(j+\epsilon,l_I+l_M-1)}\{s_{w}\}
\end{equation}
where, $\epsilon$ represents the proximity influence distance, which is used to control the continuity of the selected tokens.


\subsection{Position Encoding}
We follow the extrapolated position encoding settings as described in \citep{su2023rerope, xiao2024infllm}.
The attention weights in Equation~\ref{attention_eq} can be divided into 
(a) Long-distance attention of $C$ with respect to $I,M$: The positions of tokens in $C$ are all modified to a fixed position $w$, and the positions of tokens from $I,M$ are set to 0. The relative position will always be $w$;
and (b) Local attention of $C$ with respect to $L$ and itself: The query and key positions from $L,C$ are arranged in order of their relative positions (i.e., 0, 1, 2, ..., $l_L + l_C - 1$).
To normalize the attention weights, we fuse the two categories of attention based on their respective weights. 
The specific algorithm for computing attention is shown in Appendix~\ref{sec:pseudocode_attn}.

\subsection{Complexity Analysis} \label{Complexity_Analysis}

\paragraph{Computational Complexity Analysis.}

When inferring over a long-context sequence of total length $S$, using full-dimensional queries and keys for computing the importance score in Equation~\ref{dot-p} incurs a time complexity of $O(S^2d_H)$. By utilizing low-dimensional queries and keys, the computation is reduced to $O(S^2d^\prime)$. The additional computation for dimensionality reduction is $O(Sd_Hd^\prime+Sd^\prime)$, which scales linearly with context length and thus marginal to the quadratic savings.

In each step, ESA computes the attention matrix for a constant number of selected tokens. Considering the long sequence scenario where $M$ occupies the majority of the tokens, this approach significantly reduces computational costs, with only a minor overhead for token selection. Compared to vanilla full attention, the complexity of computing one step of attention can be reduced by ratio \( r \) in Equation~\ref{reduction_ratio_complexity}. The derivation can be found in Appendix~\ref{sec:Complexity_Analysis_Proof}.
\begin{equation} \label{reduction_ratio_complexity}
  r=\frac{2 d^{\prime} +1}{4d_{H} + 3H}
\end{equation} 

\paragraph{Cache Analysis.} 
We introduce an additional cache alongside the traditional KV cache, which stores the dimension-reduced keys from $M$. By incorporating a model that applies GQA \citep{ainslie2023gqa}, a widely used technique for reducing the KV cache, we analyze the impact of our approach on memory usage. Assuming the number of heads is denoted as \( H_G \) in GQA, the total dimensions of the kvs are given by \( d_{G} = H_G \times d \).
Given that $l_M \gg l_{I},l_{L},l_{C}$ for long sequences, we focus our analysis on the memory usage related to $M$. 
The cache size increased by the dimension-reduced keys is $\frac{d^{\prime}}{2 d_{G}}$ of the traditional KV cache.

\section{Experiments} \label{section_Experiments}
\subsection{Experimental Setup} 
\paragraph{Baselines and Benchmarks.} ESA can be applied to all decoder-only LLMs that utilize RoPE as their position encoding. We evaluate ESA using Mistral-7B-Instruct-v0.2 and Llama-3-8B-Instruct, with context lengths of 32k and 8k, respectively. We select the following three selective attention methods as our baselines: (a) InfLLM, (b) LM-Infinite (Infinite), (c) StreamingLLM (Stream). Additionally, we also choose two methods with position extrapolation: (a) NTK-aware (NTK), (b) YaRN. We conduct extensive evaluations on LongBench, $\infty$BENCH, NeedleBench, and Counting-Stars.

\paragraph{Calibration Dataset.}
We employ a small subset of Books3 data from Pile \citep{gao2020pile} as our calibration dataset to train the networks $f_{\theta_q}$ and $f_{\theta_k}$. There are 50k tokens in total and therefore 50k concatenated query-key pairs for training the networks in each layer.
The learning rate and global batch size is $0.0005$ and 128, respectively. We trained the dimensionality reduction networks for 10 epochs.

\paragraph{Parameters.}
The number of attention heads ($H$) is 32.
We compress the original size of query and key from \( d_H = 4096 \) to \( d' = 128 \).
Since the number of GQA heads is 8, the additional size required for the reduced-dimensionality keys is 6.25\% of the original KV cache.
Compared to computing full attention, the computational complexity is reduced to up to 1.56\% in each step according to Equation~\ref{reduction_ratio_complexity}.
ESA and three other baselines with selective attention select the same number of tokens. The length of initial tokens (\( l_I \)) is 128. InfLLM and ESA both choose the lengths of middle tokens and local tokens to be 2048 and 4096, respectively.
\subsection{Results on LongBench}
LongBench includes six different categories of tasks, with an average context length range from less than 10k to around 32k. We adjust the scaling factors of NTK and YaRN to accommodate the benchmark. The context length for Mistral is 32k, which does not necessitate any modification to the model's scaling factor. Consequently, we omit the NTK and YaRN for Mistral in this section. 
The results of the 16 English tasks are presented in Table~\ref{tab:res_longbench}. We draw the following conclusions: 
(a) Our method achieves improvement over the best baselines of selective attention (including Infinite, Stream, InfLLM) for both Llama and Mistral across a variety of tasks. Particularly, our method outperforms other methods of selective attention on the PassageRetrieval-en significantly, demonstrating the benefit of token-level selection.
(b) Our method is comparable to the baselines that compute full attention (including Origin, NTK, YaRN). The gap between our method and the best among these approaches is within 1 percentage point.
\input{res-table-longbench}
\subsection{Results on $\infty$BENCH} \label{results_infinitebench}
We select 6 tasks from $\infty$BENCH with an average length up to around 200k, encompassing open-form question answering (QA), code, retrieval tasks, and other domains. We set the scaling factor for NTK and YaRN to accommodate contexts of up to 200k. The results of the tasks are presented in Table~\ref{tab:res_infinitebench}. 
Firstly, our method slightly outperforms the scores of InfLLM.
The performance of our method exhibits minimal differences in retrieval tasks compared to InfLLM.
This may be due to the fact that retrieval tasks only require the model to focus on a single relevant piece of information.
InfLLM retrieves the most relevant chunks from the context, which is sufficient for solving long-text tasks that require attention to only a small amount of local information.
Our method, on the other hand, opts for retrieval at a finer granularity, resulting in performance that is close to that of InfLLM on such tasks. In other tasks, our method outperforms other selective attention methods. 
Secondly, our method outperforms NTK and YaRN, especially on Llama. This superiority may arise from the fact that methods with position embedding scaling tend to suffer a significant decline when extrapolated to excessively long contexts, such as a 8-fold extension. It demonstrates that our approach can extrapolate to significantly longer contexts, even up to a $\times 25$ extension for Llama.

\input{res-table-infbench-2col.tex}

\subsection{Results on NeedleBench and Counting-Stars}
NeedleBench and Counting-Stars evaluate the ability of LLMs to retrieve multiple pieces of related information. The two benchmarks places higher demands on the model's capacity to handle long context. Sample examples from the benchmarks are provided in Appendix~\ref{sec:NeedleBench and Counting-Stars}. The context length for these two benchmarks ranges from 4k to 256k, assessing the model's capability to retrieve multiple pieces of information across varying lengths. We uniformly set the scaling factor for NTK and YaRN to accommodate contexts of up to 200k tokens. We follow \citep{li2024needlebench, song2024counting} to use the recall accuracy as a metric to evaluate the performance of the models.

Our method exhibits great strength in extracting critical information distributed across the context.
The experimental results on Counting-Stars and NeedleBench are shown in Table~\ref{tab:res_count_stars} and ~\ref{tab:res_needle}, respectively. Details of the Counting-Stars are provided in Appendix~\ref{sec:Counting-Stars-Results}. 
Firstly, when multiple pieces of relevant information need to be retrieved, our method significantly outperforms Infinite, Stream, and InfLLM. 
This is attributed to our method's flexibility in selecting middle tokens at the token level. 
Secondly, the performance of ESA is comparable to that of NTK and YaRN. NTK and YaRN achieve promising performance by computing full attention when extrapolation is limited.
When extrapolated to tasks involving longer sequences, NTK and YaRN may experience performance degradation.
Lastly, within the original training lengths, ESA does not exhibit significant performance degradation, whereas the NTK and YaRN show a noticeable decline.
\input{res-table-counting-star-2col}
\input{res-table-needle}

\subsection{Ablation Study}

\paragraph{Effectiveness of the proximity influence distance \( \epsilon \).} The parameter \( \epsilon \) in Equation~\ref{proximity_influence_eq} controls the continuity of the selected tokens. As demonstrated in Table~\ref{tab:proximity_influence_distance_ablation}, we find this parameter to be crucial for the model, especially with regard to its retrieval capabilities. Furthermore, we observe that in Retrieve.KV, when \( \epsilon = 0,1 \), even when the model's predictions are incorrect, it is still able to retrieve parts of the correct values. For instance, the answer is "49c65968-6319-44fc-b286-feb249694b07", while the model's prediction is "49c65968-6319-44fc-\textcolor{red}{9021-cfa198896071}". 
Retrieve.KV and NeedleBench exhibit different optimal values for \( \epsilon \). 
We speculate that the underlying reason may be the difference in the number of positions where answers are to be retrieved.
In Retrieve.KV, there is typically only one segment that requires retrieval, and increasing \( \epsilon \) may enhance the completeness of the retrieved answer.
In contrast, NeedleBench involves the retrieval of answers from multiple positions. Increasing \( \epsilon \) might lead to an over-concentration of attention on a limited number of positions. 

\begin{table}[]
\centering
\resizebox{0.85\columnwidth}{!}{%
\begin{tabular}{@{}l|cccc@{}}
\toprule
\multicolumn{1}{c|}{\multirow{2}{*}{Task}} & \multicolumn{4}{c}{$\epsilon$}                \\
\multicolumn{1}{c|}{}                      & 0     & 1              & 3    & 5             \\ \midrule
Retrieve.KV                                & 66.6  & 82             & 91.6 & \textbf{95.6} \\
NeedleBench                                & 69.67 & \textbf{71.33} & 68   & 58.67         \\ \bottomrule
\end{tabular}%
}
\caption{The ablation study results of \( \epsilon \) on InfiniteBench's Retrieve.KV and NeedleBench with Mistral. NeedleBench in the table represents the average scores across lengths ranging from 4k to 200k, with the specific scores detailed in Appendix~\ref{appen_proxi_influ}.} 
  \label{tab:proximity_influence_distance_ablation}
\end{table}
  

\paragraph{Uniform Token Selection for All Heads.}
Our method does not select tokens individually for each head but rather chooses tokens based on the average importance scores across all heads. This approach is beneficial for compressing the size of queries and keys, thereby enhancing inference efficiency. To verify whether there is a significant performance degradation, we design two experiments with Llama on LongBench as shown in Table~\ref{tab:uniform_individual_token_selection}. "Individual" refers to the importance score of each token being the maximum value among the scores of all heads, meaning that each head votes for the scores. This approach ensures that the selection process takes into account all heads. "Uniform" in Table~\ref{tab:uniform_individual_token_selection} denotes our method of selecting tokens without dimensionality reduction. The scores for each subtask of LongBench are depicted in Appendix~\ref{sec:Token_Selection_for_Heads}. We extrapolate Llama from its original 8k to an average length of 32k on LongBench, and the performance on various category tasks for both token selection methods is very close. 

\begin{table}[]
\centering
\resizebox{0.85\columnwidth}{!}{%
\begin{tabular}{@{}l|cc@{}}
\toprule
\multicolumn{1}{c|}{Task} & individual & uniform \\ \midrule
LongBench scores          & 44.7       & 44.8    \\ \bottomrule
\end{tabular}%
}
\caption{We employ Llama to validate different token selection strategies for heads. The LongBench scores represent the average scores across 16 subtasks in LongBench.} 
  \label{tab:uniform_individual_token_selection}
\end{table}

\begin{figure}[t]
  \includegraphics[width=\columnwidth]{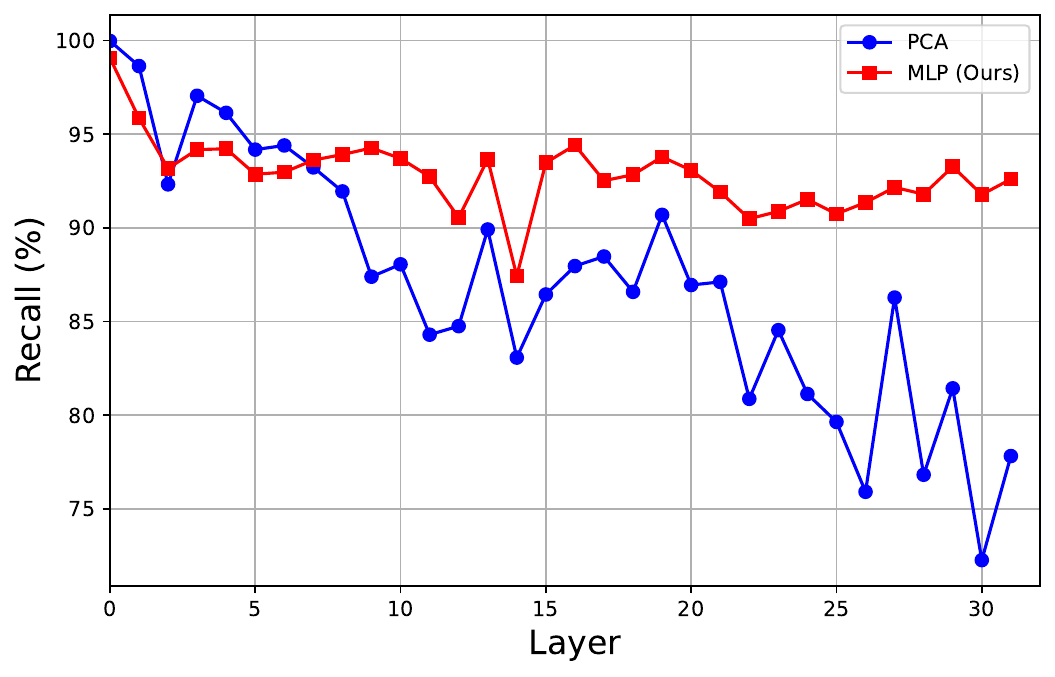}
  \caption{Recall rates of each layer for selecting the top 2,000 tokens after dimensionality reduction.}
  \label{fig:compare_pca_mlp}
\end{figure}

\paragraph{Dimension Reduction of Queries and Keys.}
We calculate the importance scores of tokens using the reduced-dimensionality queries and keys. To evaluate the impact of dimensionality reduction, we analyse experiments on LongBench with Llama using the full-dimensional query and key, as well as their reduced-dimensionality counterparts. As demonstrated in Table~\ref{tab:uniform_individual_token_selection} and Table~\ref{tab:res_longbench}, their respective scores are 44.8 and 44.41. The difference between the two scores is only 0.39.
Furthermore, we select samples from Books3 in Pile and employ Mistral to validate the recall rate of the top-k retrieval subsequent to dimensionality reduction. 
The ground truth is determined using the top-k tokens derived from the full-dimensional query and key. A total of 2,000 tokens are selected for this analysis, spanning positions from 23,000 to 25,000. 
In parallel, we execute comparative experiments utilizing principle component analysis (PCA) for dimensionality reduction inspired by \citep{singhania2024loki}. 
The experimental results are depicted in Figure~\ref{fig:compare_pca_mlp}. It can be observed that our dimensionality reduction method achieves a recall rate of over 90\% for the majority of the model's layers, whereas PCA exhibits a recall rate below 90\% in approximately half of the layers. 


\section{Conclusions and Future work}
In this paper, we propose an efficient token-level selective attention method for extending the context length of LLMs without incremental training of LLMs parameters.
The insight is to select a constant number of the most important tokens at each step for computing attention at the token level, leveraging the sparsity of the attention matrix. 
When the input sequence is sufficiently long, we are able to reduce the computational complexity to up to nearly $1.56\%$ of the original by compressing the queries and keys into low-dimensional representations. 
Our empirical evaluations demonstrate that ESA can effectively handle sequences of lengths up to $4 \times$ and even $25 \times$ the training length for various types of long sequence tasks. 
Future work could explore more accurate and efficient methods for selecting important tokens in extrapolation tasks.
\section{Limitations}
Our method has the following limitations: 1. We apply the same compression ratio to the queries and keys across different layers; employing varying compression ratios for different layers may yield better performance. 2. There may exist more effective token selection methods and compression techniques that warrant further exploration. 3. A more suitable positional encoding for our approach may exist and requires further investigation.
\bibliography{custom}

\appendix

\input{appendix_kualan}
\end{document}

%% file: intro_v2.tex
Large language models (LLMs) have demonstrated remarkable capabilities across a variety of natural language processing (NLP) tasks.
One of the emerging trends in this area is the increasing emphasis on extending the context length of LLMs to handle tasks like contextual learning and retrieval-augmented generation \citep{dong2022survey,gao2023retrieval}. However, this extension comes with its own set of challenges, particularly in terms of the Out-Of-Distribution (OOD) extrapolation on context length, massive memory demanded by key-value (KV) cache and the quadratic computational complexity of the attention mechanism. In this paper, we focus on mitigating the OOD issues and reducing the computational cost of LLM inference on long-context sequences, especially for models which adopt Rotary Position Embedding (RoPE) \citep{su2024roformer}.


Extrapolation methods based on positional embeddings address the OOD issues \citep{chen2023extending, bloc97ntk, peng2023yarn}, but they are not targeted at reducing computational cost.
To improve inference efficiency, methods focusing on \emph{selective attention} are developed, exploiting the inherent sparsity of attention matrices \citep{zhang2024h2o, Jiang:2024}. These methods usually restrict the number of selected tokens to within the pre-training context length. Early approaches focus on local attention, sometimes introduce global tokens, often neglecting long-distance tokens \citep{Beltagy2020LongformerTL,xiao2023efficient, han2023lm}. Recent chunk-based methods, such as InfLLM, LongHeads, and Landmark Attention \citep{xiao2024infllm, lu2024longheads, mohtashami2023landmark}, have aimed to extend the context length by chunking input prompts into fixed-size blocks and selecting the top-ranked chunks. 
These approaches, while reducing computational complexity, often suffer from a lack of flexibility. 
In contrast, token-level selection methods provide more flexibility by selectively attending to tokens at a finer granularity.
However, existing methods usually perform token selection only in the decoding stage and permanently evict certain tokens to manage memory usage as well as reduce computation, facing high latency during prefilling and significant information loss \citep{zhang2024h2o,liu2024scissorhands,ge2023model,li2024snapkv,Cai2024PyramidKVDK}.
To maintain accuracy while reducing computation, it is essential to perform token selection over all preceding tokens during both prefilling and decoding. Nevertheless, selecting tokens individually could introduce significant computational overhead, raising the challenge of how to efficiently identify the most crucial tokens while minimizing cost. This presents a key research problem: balancing flexibility and efficiency for selective attention approach.

To this end, we propose ESA, an efficient token-level selective attention algorithm that enables length extrapolation without the need for incremental training of LLM parameters. Specifically, our method consists of two steps: efficient selection and attention computation. 
In the first step, we introduce a query-aware token-level selection mechanism that adaptively identifies the most crucial tokens.
To reduce the computational cost, we compress the query and key vectors in the attention heads into low-dimensional representations when evaluating the importance of individual tokens. A learnable approach is employed to derive these compression functions.
Additionally, we found that directly selecting the top-ranked tokens can lead to performance degradation in certain tasks. Therefore, we propose \emph{proximity influence}, where the surrounding tokens are incorporated when calculating the importance score of a particular token.
In the step of attention computation, we utilize the full-sized keys and values of the selected tokens, rather than those of all the preceding tokens. This reduces the complexity of traditional attention computation from quadratic to linear. While there is an added computational cost for token selection, it is significantly lower than the complexity of standard attention computation.

To demonstrate the effectiveness of our approach, we conduct extensive evaluations on LongBench \citep{bai2023longbench},
$\infty$BENCH \citep{zhang2024bench}, NeedleBench \citep{li2024needlebench}, and Counting-Stars \citep{song2024counting} with Mistral-7B-Instruct-v0.2 (\texttt{Mistral}) \citep{jiang2023mistral} and Llama-3-8B-Instruct (\texttt{Llama}) \citep{llama3modelcard}.

Our contributions are summarized as follows: 1. We propose a novel token-level selective attention method for extending the context length, without incremental training of parameters of LLMs. By introducing proximity influence, we improve semantic continuity among selected tokens, addressing performance degradation caused by directly selecting top-ranked tokens in certain tasks. 2. We reduce computational complexity by leveraging low-dimensional queries and keys for token selection, achieving competitive accuracy at a significantly lower cost.


%% file: res-table-longbench.tex
\begin{table*}[]
\centering
\resizebox{\textwidth}{!}{%
\begin{tabular}{@{}l|ccccccc|ccccc@{}}
\toprule
\multicolumn{1}{c|}{\multirow{2}{*}{Task}} & \multicolumn{7}{c|}{\textbf{llama-3-8B-Instruct}}                                                                                                  & \multicolumn{5}{c}{\textbf{Mistral-7B-Instruct-v0.2}}                                                                  \\
\multicolumn{1}{c|}{}                      & Origin      & NTK(32k)    & \multicolumn{1}{c|}{YaRN(32k)}   & Infinite             & Stream               & InfLLM         & Ours                 & \multicolumn{1}{c|}{Origin}      & Infinite             & Stream         & InfLLM               & Ours                 \\ \midrule
NarrativaQA                                & 2.91        & 10.16       & \multicolumn{1}{c|}{13.14}       & 18.64                & 19.12                & 19.77          & {\ul \textbf{24.89}}       & \multicolumn{1}{c|}{20.14}       & 19.81                & 18.41          & {\ul \textbf{22.91}} & 22.72                \\
Qasper                                     & 41.44       & {\ul 44.87} & \multicolumn{1}{c|}{41.5}        & 42.35                & 42.47                & 43.51          & \textbf{43.59}       & \multicolumn{1}{c|}{29.37}       & {\ul \textbf{29.78}} & 29.74          & 28.75                & 28.65                \\
MultiFieldQA-en                            & 31.78       & {\ul 52.3}  & \multicolumn{1}{c|}{51.35}       & 38.07                & 38.41                & 44.27          & \textbf{48.23}       & \multicolumn{1}{c|}{{\ul 47.94}} & 39.03                & 38.99          & 47.54                & \textbf{47.91}       \\
MuSiQue                                    & 0.91        & 27.08       & \multicolumn{1}{c|}{{\ul 28.86}} & 19.74                & 19.89                & 22.58          & \textbf{24.23}       & \multicolumn{1}{c|}{18.56}       & 15.82                & 14.62          & 18.91                & {\ul \textbf{19.82}} \\
HotpotQA                                   & 8.24        & {\ul 53.05} & \multicolumn{1}{c|}{51.48}       & 45.4                 & 45.41                & 46.96          & \textbf{49.39}       & \multicolumn{1}{c|}{37.63}       & 31.94                & 31.57          & 36.27                & {\ul \textbf{40.06}} \\
2WikiMultihopQA                            & 30.96       & 37.51       & \multicolumn{1}{c|}{35.85}       & {\ul \textbf{39.17}} & 37.31                & 36.23          & 37.87                & \multicolumn{1}{c|}{21.61}       & 22.62                & 21.81          & 21.93                & {\ul \textbf{23.15}} \\
GovReport                                  & 18.83       & 32.48       & \multicolumn{1}{c|}{{\ul 34.22}} & 29.77                & 29.82                & \textbf{31.01} & 30.89                & \multicolumn{1}{c|}{{\ul 31.72}} & 29.52                & 29.46          & 30.97                & \textbf{31.31}       \\
QMSum                                      & 9.19        & 22.53       & \multicolumn{1}{c|}{{\ul 23.41}} & 20.92                & 20.85                & 21.37          & \textbf{22.49}       & \multicolumn{1}{c|}{{\ul 23.93}} & 21.67                & 21.77          & 23.52                & \textbf{23.79}       \\
MultiNews                                  & 26.96       & 27.46       & \multicolumn{1}{c|}{27.07}       & {\ul \textbf{27.48}} & {\ul \textbf{27.48}} & 27.33          & 27.46                & \multicolumn{1}{c|}{26.56}       & 26.32                & 26.3           & {\ul \textbf{26.63}} & 26.57                \\
TREC                                       & 52          & {\ul 75}    & \multicolumn{1}{c|}{74}          & 73                   & 73                   & 73             & \textbf{74}          & \multicolumn{1}{c|}{{\ul 71}}    & 70                   & \textbf{70.5}  & 70                   & 70                   \\
TriviaQA                                   & 30.3        & 79.39       & \multicolumn{1}{c|}{90.54}       & 90.18                & 90.34                & 90.75          & {\ul \textbf{90.91}} & \multicolumn{1}{c|}{85.81}       & 85.42                & 85.6           & 86.83                & {\ul \textbf{87.62}} \\
SAMSum                                     & 20.55       & 42.36       & \multicolumn{1}{c|}{{\ul 43.44}} & 42.12                & 42.3                 & \textbf{42.39} & 41.99                & \multicolumn{1}{c|}{{\ul 42.65}} & 41.49                & 41.69          & 42.3                 & \textbf{42.56}       \\
PassageRetrieval-en                        & 2.08        & {\ul 100}   & \multicolumn{1}{c|}{97.5}        & 39.5                 & 41                   & 70             & \textbf{86.5}        & \multicolumn{1}{c|}{{\ul 87.6}}  & 43.38                & 42.33          & 63.58                & \textbf{84.31}       \\
PassageCount                               & 2.86        & 6           & \multicolumn{1}{c|}{5.22}        & {\ul \textbf{8}}     & 7                    & 7.67           & 7.67                 & \multicolumn{1}{c|}{{\ul 3.27}}  & 1.93                 & 2.17           & 2.84                 & \textbf{3.03}        \\
LCC                                        & {\ul 59.37} & 35.43       & \multicolumn{1}{c|}{53.79}       & 58.63                & 58.94                & \textbf{59.34} & 58.58                & \multicolumn{1}{c|}{{\ul 57.15}} & 55.03                & \textbf{55.04} & 54.81                & \textbf{55.04}       \\
RepoBench-P                                & 33.92       & 33.77       & \multicolumn{1}{c|}{{\ul 53.48}} & 40.82                & 41.61                & \textbf{43.62} & 41.8                 & \multicolumn{1}{c|}{{\ul 54.55}} & 51.73                & 51.14          & 51.52                & \textbf{52.56}       \\ \midrule
Average                                    & 23.27       & 42.46       & \multicolumn{1}{c|}{{\ul 45.3}}  & 39.61                & 39.68                & 42.49          & \textbf{44.41}       & \multicolumn{1}{c|}{{\ul 41.22}} & 36.59                & 36.32          & 39.33                & \textbf{41.19}       \\ \bottomrule
\end{tabular}%
}
\caption{Results (\%) on 16 English tasks of LongBench. The term ``Origin'' indicates the original model baselines without any extrapolation methods. We \uline{underline} the best score of all methods for a model on a particular task and \textbf{bold} the best score of the selective attention methods, and this holds for the tables below.}
\label{tab:res_longbench}
\end{table*}

%% file: res-table-infbench-2col.tex
\begin{table}[]
\centering
\resizebox{\columnwidth}{!}{%
\begin{tabular}{@{}lccccccc@{}}
\toprule
\multicolumn{1}{c|}{\multirow{2}{*}{Task}} & \multicolumn{7}{c}{Method}                                                                                                \\ \cmidrule(l){2-8} 
\multicolumn{1}{c|}{}                      & Origin & NTK(200k)   & \multicolumn{1}{c|}{YaRN(200k)}  & Infinite & Stream & InfLLM               & Ours                 \\ \midrule
\multicolumn{8}{c}{\textbf{Llama-3-8B-Instruct}}                                                                                                                       \\ \midrule
\multicolumn{1}{l|}{Retrieve.KV}           & 0      & 0           & \multicolumn{1}{c|}{0}           & 1.8      & 1.8    & {\ul \textbf{4.8}}   & 3.4                  \\
\multicolumn{1}{l|}{Math.Find}             & 0      & 4.86        & \multicolumn{1}{c|}{{\ul 34.86}} & 14       & 14     & 14.86                & \textbf{16.57}       \\
\multicolumn{1}{l|}{Retrieve.Number}       & 0      & 0           & \multicolumn{1}{c|}{18.64}       & 6.44     & 6.61   & {\ul \textbf{99.66}} & 99.32                \\
\multicolumn{1}{l|}{En.MC}                 & 0      & 0           & \multicolumn{1}{c|}{19.65}       & 42.79    & 44.98  & 43.23                & {\ul \textbf{47.16}} \\
\multicolumn{1}{l|}{Code.Debug}            & 22.59  & 22.59       & \multicolumn{1}{c|}{11.17}       & 29.95    & 30.96  & 30.46                & {\ul \textbf{32.74}} \\
\multicolumn{1}{l|}{Retrieve.PassKey}      & 0      & 0           & \multicolumn{1}{c|}{51.02}       & 6.61     & 6.78   & {\ul \textbf{100}}   & {\ul \textbf{100}}   \\ \midrule
\multicolumn{1}{l|}{Average}               & 3.77   & 4.58        & \multicolumn{1}{c|}{22.56}       & 16.93    & 17.52  & 48.84                & {\ul \textbf{49.87}} \\ \midrule
\multicolumn{8}{c}{\textbf{Mistral-7B-Instruct-v0.2}}                                                                                                                  \\ \midrule
\multicolumn{1}{l|}{Retrieve.KV}           & 0      & 9.6         & \multicolumn{1}{c|}{27.8}        & 3.4      & 3.4    & {\ul \textbf{93.4}}  & 91.6                 \\
\multicolumn{1}{l|}{Math.Find}             & 27.43  & {\ul 29.43} & \multicolumn{1}{c|}{25.14}       & 14       & 14.29  & \textbf{26.86}       & \textbf{26.86}       \\
\multicolumn{1}{l|}{Retrieve.Number}       & 27.29  & 90.34       & \multicolumn{1}{c|}{96.95}       & 6.78     & 6.78   & {\ul \textbf{99.83}} & {\ul \textbf{99.83}} \\
\multicolumn{1}{l|}{En.MC}                 & 10.4   & 18.78       & \multicolumn{1}{c|}{40.61}       & 38.43    & 37.99  & 43.23                & {\ul \textbf{48.03}} \\
\multicolumn{1}{l|}{Code.Debug}            & 15.51  & 20.81       & \multicolumn{1}{c|}{21.83}       & 22.84    & 22.59  & 27.41                & {\ul \textbf{34.01}} \\
\multicolumn{1}{l|}{Retrieve.PassKey}      & 75.25  & {\ul 100}         & \multicolumn{1}{c|}{{\ul 100}}         & 6.78     & 6.78   & {\ul \textbf{100}}   & {\ul \textbf{100}}   \\ \midrule
\multicolumn{1}{l|}{Average}               & 36.31  & 44.83       & \multicolumn{1}{c|}{52.06}       & 15.37    & 15.31  & 65.12                & {\ul \textbf{66.72}} \\ \bottomrule
\end{tabular}%
}
\caption{Performace evaluation (\%) on 6 English tasks of $\infty$BENCH.}
\label{tab:res_infinitebench}
\end{table}

%% file: res-table-counting-star-2col.tex
\begin{table}[]
\centering
\resizebox{\columnwidth}{!}{%
\begin{tabular}{@{}lccccccc@{}}
\toprule
\multicolumn{1}{c|}{\multirow{2}{*}{Task}} & \multicolumn{7}{c}{Method}                                                                              \\ \cmidrule(l){2-8} 
\multicolumn{1}{c|}{}                      & Origin & NTK(200k) & \multicolumn{1}{c|}{YaRN(200k)} & Infinite & Stream & InfLLM & Ours                \\ \midrule
\multicolumn{8}{c}{\textbf{Llama-3-8B-Instruct}}                                                                                                     \\ \midrule
\multicolumn{1}{l|}{(128k, 32, 32)}        & 5      & 28.5      & \multicolumn{1}{c|}{8.4}        & 15.8     & 15.7   & 24.2   & {\ul \textbf{39.3}} \\
\multicolumn{1}{l|}{(256k, 8, 32)}         & 2.7    & 16        & \multicolumn{1}{c|}{5.9}        & 5.1      & 5.1    & 10.9   & {\ul \textbf{63.7}} \\
\multicolumn{1}{l|}{(256k, 16, 32)}        & 2      & 16        & \multicolumn{1}{c|}{8.2}        & 11.7     & 11.1   & 14.1   & {\ul \textbf{44.3}} \\
\multicolumn{1}{l|}{(256k, 32, 32)}        & 1.7    & 15.7      & \multicolumn{1}{c|}{7.7}        & 9.6      & 9.6    & 16.2   & {\ul \textbf{36.4}} \\ \midrule
\multicolumn{1}{l|}{Average}               & 2.9    & 19.1      & \multicolumn{1}{c|}{7.6}        & 10.6     & 10.4   & 16.4   & {\ul \textbf{45.9}} \\ \midrule
\multicolumn{8}{c}{\textbf{Mistral-7B-Instruct-v0.2}}                                                                                                \\ \midrule
\multicolumn{1}{l|}{(128k, 32, 32)}        & 15     & 35.5      & \multicolumn{1}{c|}{{\ul 52.6}} & 10.7     & 10.2   & 16.1   & \textbf{26.6}       \\
\multicolumn{1}{l|}{(256k, 8, 32)}         & 5.5    & 6.6       & \multicolumn{1}{c|}{{\ul 46.5}} & 5.9      & 7      & 2.7    & \textbf{25.8}       \\
\multicolumn{1}{l|}{(256k, 16, 32)}        & 6.2    & 15.2      & \multicolumn{1}{c|}{{\ul 47.3}} & 9        & 8.6    & 12.7   & \textbf{16.8}       \\
\multicolumn{1}{l|}{(256k, 32, 32)}        & 9.8    & 16.1      & \multicolumn{1}{c|}{{\ul 41.7}} & 10.9     & 10.9   & 13.3   & \textbf{23.2}       \\ \midrule
\multicolumn{1}{l|}{Average}               & 9.1    & 18.4      & \multicolumn{1}{c|}{{\ul 47}}   & 9.1      & 9.2    & 11.2   & \textbf{23.1}       \\ \bottomrule
\end{tabular}%
}
\caption{Recall accuracy (\%) evaluation on the Counting-Stars benchmark. We employ the notation (256k, 8, 32) to represent the following benchmark setup: the context length is 256k, and within this length, we generate 32 samples at equal intervals (e.g., the sample lengths are 8k, 16k, 32k, ..., up to 256k), with each sample containing 8 pieces of relevant information. Accuracy is defined as the average score across the 32 samples within each task.}
\label{tab:res_count_stars}
\end{table}

%% file: res-table-needle.tex
\begin{table*}[]
\centering
\resizebox{\textwidth}{!}{%
\begin{tabular}{@{}l|ccccccc|ccccccc@{}}
\toprule
\multirow{2}{*}{\begin{tabular}[c]{@{}l@{}}Context\\ Length\end{tabular}} & \multicolumn{7}{c|}{\textbf{Llama-3-8B-Instruct}}                                                                                                       & \multicolumn{7}{c}{\textbf{Mistral-7B-Instruct-v0.2}}                                                                                                 \\
                                                                          & Origin      & NTK(200k)   & \multicolumn{1}{c|}{YaRN(200k)} & Infinite             & Stream               & InfLLM               & Ours                 & Origin      & NTK(200k)   & \multicolumn{1}{c|}{YaRN(200k)}     & Infinite           & Stream             & InfLLM             & Ours                 \\ \midrule
4k                                                                        & {\ul 96.67} & 90          & \multicolumn{1}{c|}{70}         & {\ul \textbf{96.67}} & {\ul \textbf{96.67}} & {\ul \textbf{96.67}} & {\ul \textbf{96.67}} & 96.67       & 76.67       & \multicolumn{1}{c|}{60}             & {\ul \textbf{100}} & {\ul \textbf{100}} & {\ul \textbf{100}} & {\ul \textbf{100}}   \\
8k                                                                        & {\ul 100}   & {\ul 100}   & \multicolumn{1}{c|}{76.67}      & 73.33                & 73.33                & 86.67                & {\ul \textbf{100}}   & 80          & 73.33       & \multicolumn{1}{c|}{43.33}          & 50                 & 53.33              & 63.33              & {\ul \textbf{83.33}} \\
16k                                                                       & 0           & {\ul 96.67} & \multicolumn{1}{c|}{43.33}      & 20                   & 20                   & 23.33                & \textbf{86.67}       & 73.33       & {\ul 76.67} & \multicolumn{1}{c|}{66.67}          & 13.33              & 6.67               & 10                 & \textbf{73.33}       \\
48k                                                                       & 0           & 36.67       & \multicolumn{1}{c|}{0}          & 3.33                 & 3.33                 & 3.33                 & {\ul \textbf{53.33}} & {\ul 83.33} & 50          & \multicolumn{1}{c|}{53.33}          & 0                  & 0                  & 0                  & \textbf{70}          \\
80k                                                                       & 0           & 0           & \multicolumn{1}{c|}{0}          & 0                    & 0                    & 0                    & {\ul \textbf{36.67}} & 6.67        & 63.33       & \multicolumn{1}{c|}{{\ul 66.67}}    & 0                  & 0                  & 0                  & {\ul \textbf{66.67}} \\
112k                                                                      & 0           & 0           & \multicolumn{1}{c|}{0}          & 0                    & 0                    & 0                    & {\ul \textbf{43.33}} & 0           & {\ul 73.33} & \multicolumn{1}{c|}{70}             & 0                  & 0                  & 0                  & \textbf{63.33}       \\
128k                                                                      & 0           & 0           & \multicolumn{1}{c|}{0}          & 0                    & 0                    & 0                    & {\ul \textbf{46.67}} & 0           & {\ul 73.33} & \multicolumn{1}{c|}{63.33}          & 0                  & 0                  & 0                  & \textbf{60}          \\
144k                                                                      & 0           & 0           & \multicolumn{1}{c|}{0}          & 0                    & 0                    & 0                    & {\ul \textbf{20}}    & 0           & 43.33       & \multicolumn{1}{c|}{{\ul 66.67}}    & 0                  & 0                  & 0                  & \textbf{63.33}       \\
176k                                                                      & 0           & 0           & \multicolumn{1}{c|}{0}          & 0                    & 0                    & 0                    & {\ul \textbf{23.33}} & 0           & 0           & \multicolumn{1}{c|}{{\ul 70}}       & 0                  & 0                  & 0                  & \textbf{53.33}       \\
200k                                                                      & 0           & 0           & \multicolumn{1}{c|}{0}          & 0                    & 0                    & 0                    & {\ul \textbf{23.33}} & 0           & 0           & \multicolumn{1}{c|}{{\ul 66.67}}    & 0                  & 0                  & 0                  & \textbf{46.67}       \\ \midrule
Average                                                                   & 19.67       & 32.33       & \multicolumn{1}{c|}{19}         & 19.33                & 19.33                & 21                   & {\ul \textbf{53}}    & 34          & 53          & \multicolumn{1}{c|}{\textbf{62.67}} & 16.33              & 16                 & 17.33              & {\ul \textbf{68}}    \\ \bottomrule
\end{tabular}%
}
  \caption{Recall accuracy (\%) evaluation on NeedleBench. 
  } 
  \label{tab:res_needle}
\end{table*}

%% file: appendix_kualan.tex
\onecolumn

\section{Complexity Analysis Derivation}
\label{sec:Complexity_Analysis_Proof}

Denote $l_{I,M,L,C} = l_{I} + l_{M} + l_{L} + l_{C}$, the full attention in each step is computed as
\begin{equation} \label{attention_eq_full}
	\mathbf{O} = \text{Attn}(\mathbf{Q}_C,\mathbf{K}_{[I,M,L,C]}, \mathbf{V}_{[I,M,L,C]})
\end{equation}
The complexity of computing full attention consists of the following parts: 
\begin{enumerate}
\item The dot product of queries and keys: 
\begin{equation} \label{attn_comple_dot}
  2 \cdot d_{H} \cdot l_{C} \cdot l_{I,M,L,C} 
\end{equation}
\item The softmax operation (including exponential calculation, summation, and normalization): 
\begin{equation} \label{attn_comple_softmax}
  3 \cdot H \cdot l_{C} \cdot l_{I,M,L,C}
\end{equation}
\item The dot product of attention weights and values: 
\begin{equation} \label{attn_comple_wei_val}
  2 \cdot d_{H} \cdot l_{C} \cdot l_{I,M,L,C}
\end{equation}
\end{enumerate}

The overall complexity is
\begin{equation} \label{attn_full}
  (4 \cdot d_{H} + 3 \cdot H) \cdot l_{C} \cdot l_{I,M,L,C} 
\end{equation}

The complexity of our method comprises the following components:
\begin{enumerate}
    \item Reduction of the dimensions of the query and key in Equation~(\ref{reduce_dim_formula}):
    \begin{equation}
      2 \cdot 2 \cdot l_C \cdot d_{H} \cdot d' + 2 \cdot l_C \cdot d'
    \end{equation}
    \item The complexity of token selection in Equation~(\ref{F_s_1}) and~(\ref{proximity_influence_eq}) is
    $2 \cdot l_M \cdot l_C \cdot d' + l_{M} \cdot l_C + h_{\text{max}}(2 \cdot l_C \cdot l_M + (1 + 2 \epsilon) \cdot l_M)$. The complexity of taking the maximum operation on multi-dimensional vectors is denoted as \( h_{\text{max}} \). Since (a) the complexity is lower than that of an equivalent number of Floating Point Operations (FLOPS)  for the same scale; and (b) $2 \cdot l_C \cdot l_M + (1 + 2 \epsilon) \cdot l_M < 2 \cdot l_M \cdot l_C \cdot d'$, we neglect the impact of the max operation. Therefore, the complexity of token selection is:
    \begin{equation}
      2 \cdot l_M \cdot l_C \cdot d' + l_{M} \cdot l_C
    \end{equation}
    \item Following the analogy of Equation~(\ref{attn_comple_dot}),~(\ref{attn_comple_softmax}) and~(\ref{attn_comple_wei_val}), after selecting $k$ tokens from $M$, the complexity of computing sparse attention is as follows:
    \begin{equation}
      (4 \cdot d_{H} + 3 \cdot H ) \cdot l_{C} \cdot (l_{I,L,C}+k)
    \end{equation}
    where, $l_{I,L,C} \triangleq l_{I}+l_{L}+l_{C}$.
\end{enumerate}
Given that $l_M >> l_{I},l_{L},l_{C},k,d_{H},d^{\prime}$, the overall reduction ratio of complexity is:
\begin{align} \label{reduction_ratio_complexity_infer}
r = \frac{l_I + k + l_L + l_C}{l_I + l_M + l_L + l_C} + \frac{4 d_{H} d^{\prime} + 2 d^{\prime} + 2 d^{\prime} l_M + l_M}{(4 d_{H} + 3H) (l_I + l_M + l_L + l_C)} \approx \frac{2 d^{\prime} +1}{4d_{H} + 3H}
\end{align}

\section{Model Loading}
\label{sec:load_model_appendix}
All of our experiments were conducted on a device equipped with 4 $\times$ A800 GPUs. The evaluated models were partitioned across the 4 GPUs by layer using the \texttt{accelerate} \citep{accelerate}. Our model loading approach is capable of supporting the execution of all the aforementioned experiments. 
Additionally, we support loading the model onto a single A800 GPU by offloading the original KV cache to the CPU, while the dimension-reduced keys are always kept on the GPU. 
During each attention computation, a small number of keys and values in KV cache are loaded onto the GPU. 
By employing this cache management approach, we are able to perform inference on long sequence tasks with lengths of 700k+ on a single A800 GPU.
For NTK and YaRN, we employ vLLM \citep{kwon2023efficient} for inference on a device equipped with 4 $\times$ A800 GPUs.

\section{Samples of NeedleBench and Counting-Stars}
\label{sec:NeedleBench and Counting-Stars}

\paragraph{NeedleBench}
A sample from NeedleBench is shown below, where \textit{xxxxxxxx} represents noise text:
\begin{quote}
  You are an intelligent AI assistant skilled in answering user questions. Please keep your answers concise and clear. Do not talk about irrelevant topics or repeat your answers.\\
  The document given to you by the user is May 2001 xxxxxxxx \textbf{Hidden on Hell Island is the legendary Dream Bubble.}  xxxxxxxx \textbf{Hidden on Emerald Island is the legendary Ghost Pearl.}   xxxxxxxx \textbf{Hidden on Sand Island is the legendary Stardust Shard.}xxxxxxxx\\
  Now, the questions are: \textbf{What legendary item is hidden on Hell Island?What legendary item is hidden on Emerald Island?What legendary item is hidden on Sand Island?}Before answering, please consider what in the document is most relevant to this question. Please answer in the format of 'The legendary item hidden on the Hell Island is\underline{$\phantom{\rule{1cm}{0ex}}$}. The legendary item hidden on the Emerald Island is\underline{$\phantom{\rule{1cm}{0ex}}$}. The legendary item hidden on the Sand Island is\underline{$\phantom{\rule{1cm}{0ex}}$}
\end{quote}

\paragraph{Counting-Stars}
A sample from Counting-Stars is shown below, where \textit{xxxxxxxx} represents noise text:
\begin{quote}
  xxxxxxxx \textbf{The little penguin counted 15 $\star$} xxxxxxxx \textbf{The little penguin counted 117 $\star$} xxxxxxxx \textbf{The little penguin counted 42 $\star$} xxxxxxxx \textbf{The little penguin counted 29 $\star$} \\
  On this moonlit and misty night, the little penguin is looking up at the sky and concentrating on counting $\star$. Please help the little penguin collect the number of $\star$, for example: \{"little\_penguin": [x, x, x,...]\}. The summation is not required, and the numbers in [x, x, x,...] represent the counted number of $\star$ by the little penguin. Only output the results in JSON format without any explanation. 
  ```json \\
  \{"little\_penguin": [
\end{quote}

\section{Proximity Influence Distance: Specific Experiments} \label{appen_proxi_influ}
We validate the ablation study results of the proximity influence distance, with the detailed findings presented in Table~\ref{tab:proximity_influence_distance_details}. 
When \( \epsilon \) is set to 0 and 1, Mistral demonstrates superior performance across all subtasks ranging from 4k to 200k. 
As \( \epsilon \) is further increased to 3 and 5, the scores on the short-sequence subtasks (4k-16k) remain comparable to the previous results. 
However, the model's performance exhibits a significant decline on the long-sequence subtasks (48k-200k).

\begin{table}[]
\centering
\resizebox{\columnwidth}{!}{%
\begin{tabular}{c|cccccccccc|c}
\hline
\multicolumn{1}{l|}{\multirow{2}{*}{$\epsilon$}} & \multicolumn{10}{c|}{Context Length} & \multicolumn{1}{l}{\multirow{2}{*}{Average}} \\
\multicolumn{1}{l|}{} & \multicolumn{1}{l}{4k} & \multicolumn{1}{l}{8k} & \multicolumn{1}{l}{16k} & \multicolumn{1}{l}{48k} & \multicolumn{1}{l}{80k} & \multicolumn{1}{l}{112k} & \multicolumn{1}{l}{128k} & \multicolumn{1}{l}{144k} & \multicolumn{1}{l}{176k} & \multicolumn{1}{l|}{200k} & \multicolumn{1}{l}{} \\ \hline
0 & 100 & 83.33 & \textbf{73.33} & 80 & \textbf{70} & 60 & 66.67 & \textbf{70} & 46.67 & \textbf{46.67} & 69.67 \\
1 & 100 & 83.33 & \textbf{73.33} & \textbf{86.67} & 63.33 & \textbf{70} & \textbf{73.33} & 56.67 & \textbf{60} & \textbf{46.67} & \textbf{71.33} \\
3 & 100 & 83.33 & \textbf{73.33} & 70 & 66.67 & 63.33 & 60 & 63.33 & 53.33 & \textbf{46.67} & 68 \\
5 & 100 & 83.33 & 70 & 70 & 40 & 40 & 56.67 & 36.67 & 50 & 40 & 58.67 \\ \hline
\end{tabular}%
}
\caption{We validate the performance across all subtasks of NeedleBench with varying \( \epsilon \) using Mistral.}
  \label{tab:proximity_influence_distance_details}
\end{table}

\section{Token Selection for Heads: Specific Experiments}
\label{sec:Token_Selection_for_Heads}
We employ Llama to investigate the impact of different token selection methods for heads. 
We extrapolate Llama's original context length of 8k to 32k and conduct experiments using LongBench, which has an average length of 32k.
The scores for each subtask are presented in Table~\ref{tab:token_selection_head_subtask}.
\begin{table}[]
\centering
  \begin{tabular}{l|cc}
  \hline
  \multicolumn{1}{c|}{Task} & individual     & uniform        \\ \hline
  NarrativaQA               & 24.52          & \textbf{25.1}  \\
  Qasper                    & \textbf{44.69} & 44.51          \\
  MultiFieldQA-en           & 49.18          & 49.18          \\
  MuSiQue                   & 25.55          & \textbf{27.58} \\
  HotpotQA                  & \textbf{49.57} & 49.26          \\
  2WikiMultihopQA           & \textbf{38.1}  & 37.44          \\
  GovReport                 & \textbf{31.06} & 30.99          \\
  QMSum                     & \textbf{22.91} & 22.75          \\
  MultiNews                 & 27.41          & \textbf{27.45} \\
  TREC                      & 73.5           & 73.5           \\
  TriviaQA                  & 91.19          & 91.19          \\
  SAMSum                    & \textbf{42.87} & 42.7           \\
  PassageRetrieval-en       & 86.5           & \textbf{87.5}  \\
  PassageCount              & 8.17           & 7.17           \\
  LCC                       & 58.32          & 58.32          \\
  RepoBench-P               & 41.7           & \textbf{42.6}  \\ \hline
  Average                   & 44.7           & \textbf{44.8}           \\ \hline
  \end{tabular}
  \caption{Llama's specific experiments on LongBench using different token selection methods for heads. "Individual" refers to the importance score of each token being the maximum value among the scores of all heads, meaning that each head votes for the scores. This approach ensures that the selection process takes into account all heads. "Uniform" in the table denotes our method of selecting tokens without dimensionality reduction.}
  \label{tab:token_selection_head_subtask}
  \end{table}

\section{LM-Infinite with additional top-k middle tokens}
 LM-Infinite optionally select top-k middle tokens for some higher layers for each head. Following the settings in their paper, we evaluate the effectiveness of this method and the results are demonstrated in Table~\ref{tab:infinite-top}. The performance is improved compared to the original LM-Infinite setting. Nevertheless, it is not as competitive as our ESA method.
\begin{table}[]
    \centering
    \begin{tabular}{@{}l|c|c@{}}
        \toprule
        \multicolumn{1}{c|}{Task} & Llama & Mistral \\ \midrule
        NarrativaQA               & 20.6  & 22.02   \\
        Qasper                    & 21.73 & 30.36   \\
        MultiFieldQA-en           & 40.2  & 44.52   \\
        MuSiQue                   & 20.19 & 16.36   \\
        HotpotQA                  & 44.66 & 32.63   \\
        2WikiMultihopQA           & 38.09 & 22.64   \\
        GovReport                 & 31.28 & 31.61   \\
        QMSum                     & 21.73 & 22.5    \\
        MultiNews                 & 27.54 & 26.7    \\
        TREC                      & 73.5  & 70.5    \\
        TriviaQA                  & 90.91 & 86.59   \\
        SAMSum                    & 42.57 & 42.26   \\
        PassageRetrieval-en       & 38.5  & 49.42   \\
        PassageCount              & 8.5   & 2.37    \\
        LCC                       & 60.75 & 57.4    \\
        RepoBench-P               & 43.83 & 53.51   \\ \midrule
        Average                   & 40.37 & 38.212  \\ \bottomrule
    \end{tabular}
    \caption{Results on LongBench with Infinite-LM attending to top 5 tokens in the middle.}
    \label{tab:infinite-top}
\end{table}

\section{Pseudocode for Computing Attention}
\label{sec:pseudocode_attn}
We support the computation of local and global attention using either Flash Attention \citep{dao2205fast} or PyTorch operators. The pseudocode for computing a step of attention with Flash Attention is shown in Algorithm~\ref{alg:Pseudocode for Attention Computation}. It is worth noting that we omit the exp-normalize trick in Step 12 of Algorithm~\ref{alg:Pseudocode for Attention Computation} to avoid numerical overflow. We employ the function \texttt{flash\_attn\_func} provided by Flash Attention, which returns the logarithm of the softmax normalization factor as its second result. In environments where Flash Attention is not supported, we can replace the function \texttt{flash\_attn\_func} with PyTorch operators.
\begin{algorithm*} 
\caption{Pseudocode for Attention Computation with Flash Attention}
\begin{algorithmic}[1]
\State \textbf{Input:}
\State $l\_q$: Queries from $C$ with position encoding $(l_L, l_L+1, l_L+2, \ldots, l_L + l_C - 1)$
\State $g\_q$: Queries from $C$ with position encoding $(w, w, w, \ldots, w)$
\State $l\_k$: Concatenated keys from $L$ and $C$ with position encoding $(0, 1, 2, \ldots, l_L + l_C - 1)$
\State $g\_k$: Selected keys from $M$ and $I$ with position encoding $(0, 0, 0, 0, \ldots)$
\State $l\_v$: Concatenated values from $L$ and $C$
\State $g\_v$: Selected values from $M$ and $I$
\State
\State \textbf{Procedure:}
\State $(l\_attn, l\_lse, \_) \leftarrow \text{flash\_attn\_func}(l\_q, l\_k, l\_v, \mathrm{causal}=\mathrm{True})$
\State $(g\_attn, g\_lse, \_) \leftarrow \text{flash\_attn\_func}(g\_q, g\_k, g\_v, \mathrm{causal}=\mathrm{False})$
\State $se \leftarrow \exp([l\_lse, g\_lse])$
\State $fac \leftarrow se / \sum se$
\State $attn \leftarrow [l\_attn, g\_attn] \cdot fac$
\State
\State \textbf{Output:}
\State $attn$
\end{algorithmic}
\label{alg:Pseudocode for Attention Computation}
\end{algorithm*}
